\documentclass{article}
\usepackage{spconf,amsmath,graphicx,hyperref,shared_figstyles,algorithm,algpseudocode}
\usepackage{booktabs}
\title{Enhancing Speech Emotion Recognition via Fine-Tuning Pre-Trained Models and Hyper-Parameter Optimisation}
\name{Aryan Golbaghi,  Shuo Zhou}
\address{School of Computer Science, University of Sheffield, Sheffield, UK\\
\small{\tt \{agolbaghi1, shuo.zhou\}@sheffield.ac.uk}}
\begin{document}
%\ninept
%
\maketitle
\begin{abstract}
We propose a workflow for speech emotion recognition (SER) that combines pre-trained representations with automated hyperparameter optimisation (HPO). Using SpeechBrain’s wav2vec2-base model fine-tuned on IEMOCAP as the encoder, we compare two HPO strategies, Gaussian Process Bayesian Optimisation (GP-BO) and Tree-structured Parzen Estimators (TPE), under an identical four-dimensional search space and 15-trial budget, with balanced class accuracy (BCA) on the German EmoDB corpus as the objective. All experiments run on 8 CPU cores with 32 GB RAM. GP-BO achieves 96.0\% BCA in 11 minutes, and TPE (Hyperopt implementation) attains 97.0\% in 15 minutes. In contrast, grid search requires 143 trials and 1,680 minutes to exceed 90\% BCA, and the best AutoSpeech 2020 baseline reports only 85\% in 30 minutes on GPU. For cross-lingual generalisation, an EmoDB-trained HPO-tuned model improves zero-shot accuracy by 25\% on CREMA-D and 26\% on RAVDESS. Results show that efficient HPO with pre-trained encoders delivers competitive SER on commodity CPUs. Source code to this work is available at: \url{https://github.com/youngaryan/speechbrain-emotion-hpo}.
\end{abstract}

\begin{keywords}
speech emotion recognition, pre-trained model, hyperparameter optimisation
\end{keywords}
\section{Introduction}
\label{sec:intro}

Speech emotion recognition (SER) plays a key role in human-computer interaction, mental health monitoring, and social robotics \cite{el2011survey, schuller2018speech}. Strong accuracy has typically relied on GPU resources and extensive manual tuning of hyperparameters. We target a low-cost, CPU-only pipeline that remains competitive with recent SER systems while requiring minimal expert intervention. To our knowledge, this is the first SER approach that combines a self-supervised backbone with automated machine learning (AutoML) to achieve state-of-the-art performance on commodity hardware.  

Early SER systems relied on handcrafted acoustic and prosodic features such as MFCCs, pitch, and energy, paired with classical classifiers \cite{schuller2009interspeech, eyben2015geneva}. In the 2010s, deep neural networks, including CNNs, RNNs, and hybrid architectures, became dominant by learning hierarchical time–frequency patterns directly from spectrograms \cite{han2014speech, trigeorgis2016adieu}. More recently, self-supervised pre-training has enabled general-purpose audio representations that can be fine-tuned for SER, with notable examples including wav2vec 2.0 \cite{baevski2020wav2vec}, HuBERT \cite{hsu2021hubert}, and ExHuBERT \cite{amiriparian24_interspeech}.  

In parallel, automated machine learning (AutoML) techniques including hyperparameter optimisation (HPO)  such as Gaussain Process Bayesian optimisation (GP-BO) \cite{snoek2012practical} and tree-structured Parzen estimators (TPE) \cite{bergstra2011algorithms} have reduced the burden of manual tuning in many domains, but remain underexplored in SER. The AutoSpeech 2020 challenge \cite{wang2020autospeech} demonstrated the potential of automated pipelines for speech classification, though its reliance on GPU resources limited accessibility for low-cost or CPU-only deployments.  

Despite these advances, three key challenges persist: (i) the high computational cost of model selection, (ii) the difficulty of optimising many interdependent hyperparameters, and (iii) limited cross-corpus and cross-language generalisability \cite{schuller2011recognising, wani2021comprehensive}. HPO methods such as GP-BO and TPE offer a principled way to mitigate these issues by reducing manual search effort while preserving competitive accuracy.

\section{Architecture}
\label{sec:architecture}
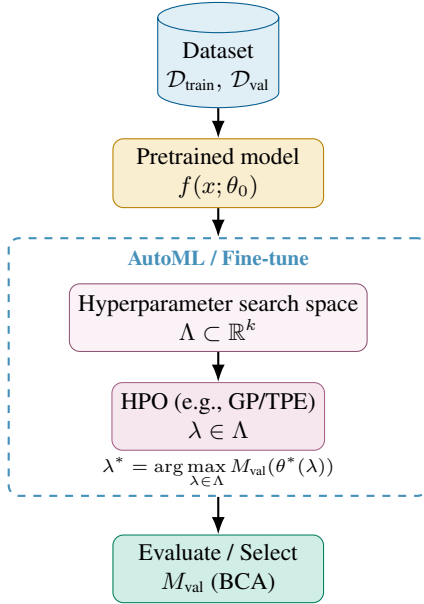
\begin{figure}[t]
  \centering
  \begin{tikzpicture}[
  % node distance and styles come from shared_figstyles.sty
]
% ---- Main flow (top) -------------------------------------------------------
\node[dataset]  (emodb)    {Dataset\\ $\mathcal{D}_{\text{train}},\,\mathcal{D}_{\text{val}}$};
\node[model]    (pretrain) [below=4mm of emodb]   {Pretrained model\\ $f(x;\theta_0)$};

% ---- Place both inner nodes BELOW the pretrained model ---------------------
\node[space]    (space)    [below=10mm of pretrain] {Hyperparameter search space\\ $\Lambda\subset\mathbb{R}^k$};
\node[hpo]      (optuna)   [below=4mm  of space]    {HPO (e.g., GP/TPE)\\ $\lambda\in\Lambda$};

% Fine-tune group around the HPO stack
\node[
  fintuneBox,
  fit=(space)(optuna),
  inner xsep=8mm,
  inner ysep=6mm,
  label={[font=\footnotesize\bfseries, text=oiSky!80!black,
          label distance=-5mm]north:AutoML / Fine-tune}
] (finbox) {};

% Evaluate/Select below the Fine-tune group
\node[metrics]  (metrics)  [below=5mm of finbox]  {Evaluate / Select\\ $M_{\text{val}}$ (BCA)};

% ---- Arrows ----------------------------------------------------------------
\draw[arrow] (emodb)        -- (pretrain);
\draw[arrow] (pretrain)     -- (finbox.north);
\draw[arrow] (finbox.south) -- (metrics.north);
\draw[arrow] (space)        -- (optuna);

\node[font=\scriptsize, align=center, inner sep=5pt,
      anchor=north, yshift=1mm] at (optuna.south)
  {$\displaystyle \lambda^*=\arg\max_{\lambda\in\Lambda} M_{\text{val}}(\theta^*(\lambda))$};
\end{tikzpicture}
  \caption{Proposed AutoML fine-tuning workflow for SER. A pre-trained model is fine-tuned on $\mathcal{D}_{\text{train}}$, with HPO guiding hyperparameter search and $M_{\text{val}}$ used for model selection.}
  \label{fig:automl-hpo}
\end{figure}

We propose an AutoML fine-tuning workflow for SER, as demonstrated in Fig. \ref{fig:automl-hpo}. The inputs are: (i) a dataset split into training and validation sets, (ii) a pre-trained speech model, (iii) a $k$-dimensional hyperparameter vector, and (iv) an HPO algorithm.  

\subsection{Problem formulation}

Let $\mathcal{D}_{\text{train}} = \{(x_i, y_i)\}_{i=1}^{N_{\text{tr}}}$ and $\mathcal{D}_{\text{val}} = \{(x_j, y_j)\}_{j=1}^{N_{\text{val}}}$ denote training and validation sets. The encoder is initialised from a pre-trained network $f(\cdot; \theta_0)$. A hyperparameter vector $\lambda \in \Lambda \subset \mathbb{R}^k$ controls fine-tuning parameters such as learning rate, weight decay, scheduler settings, batch size, and dropout. For a given $\lambda$, parameters are obtained as
\begin{equation}
\theta^\star(\lambda) \leftarrow \text{Train}(f(\cdot;\theta_0), \lambda; \mathcal{D}_{\text{train}}).
\end{equation}
Performance is measured by a validation metric $M_{\text{val}}$ (balanced class accuracy, BCA), and the optimal configuration is selected as
\begin{equation}
\lambda^\star = \arg\max_{\lambda \in \Lambda} M_{\text{val}}(\theta^\star(\lambda)).
\end{equation}
For a classification problem with $C$ classes, BCA is 
\begin{equation}
\text{BCA} = \frac{1}{C}\sum_{c=1}^{C} \frac{\text{TP}_c}{\text{TP}_c + \text{FN}_c},
\end{equation}
where $\text{TP}_c$ and $\text{FN}_c$ are true positives and false negatives for class $c$.  

\subsection{HPO algorithms}
We incorporate two families of sequential model-based optimisation (SMBO, Algorithm~\ref{alg:smbo}) as HPO candidates in our workflow and compare their performance
(i) \textbf{Gaussian process Bayesian optimisation (GP-BO)} \cite{snoek2012practical}, and  
(ii) \textbf{Tree-structured Parzen estimators (TPE)} \cite{bergstra2011algorithms}.

At iteration $t$, a surrogate is fit on history $\mathcal{D}_{t-1}$, candidate hyperparameters $\lambda_t$ are selected by maximising an acquisition function, and the objective is evaluated as $y_t = f(\lambda_t)$. The history is then updated, and the process repeats.  

\paragraph*{GP-BO.}  
Assume noisy scores $y = f(\lambda) + \epsilon$ with $\epsilon \sim \mathcal{N}(0,\sigma_n^2)$ and a GP prior on $f$. Conditioned on $\mathcal{D}_{t-1} = \{(\lambda_i,y_i)\}_{i=1}^{t-1}$, the predictive distribution is  
\begin{equation}
f(\lambda) \mid \mathcal{D}_{t-1} \sim \mathcal{N}(\mu_t(\lambda), \sigma_t^2(\lambda)),
\end{equation}
where $\mu_t$ is the GP mean and $\sigma_t$ the predictive variance. Acquisition functions such as Expected Improvement (EI) \cite{jones1998efficient} or GP-UCB \cite{srinivas2010gaussian} balance exploration and exploitation.  

\paragraph*{TPE.}  
Instead of modelling $p(y|\lambda)$ directly, TPE estimates densities over configurations:  
\begin{equation}
\ell(\lambda) = p(\lambda \mid y < y^\star), \quad g(\lambda) = p(\lambda \mid y \geq y^\star),
\end{equation}
where $y^\star$ is a quantile threshold. Candidates are chosen by maximising $\ell(\lambda)/g(\lambda)$, which is proportional to expected improvement under this model.

\begin{algorithm}[t]
\caption{Sequential Model-Based Optimisation (SMBO)}
\label{alg:smbo}
\begin{algorithmic}[1]
\Require Search space $\Lambda$, budget $T$, initial design size $n_0$, acquisition $\alpha(\cdot)$
\State \textbf{Initialize} $\mathcal D_0 \leftarrow \{(\lambda_i, y_i)\}_{i=1}^{n_0}$ by evaluating $y_i = f(\lambda_i)+\varepsilon_i$ on a diverse design over $\Lambda$
\State $\lambda^\star \leftarrow \arg\min_{(\lambda,y)\in\mathcal D_0} y$ \Comment{incumbent}
\For{$t = 1$ to $T$}
  \State Fit surrogate $s_t$ on $\mathcal D_{t-1}$ \Comment{e.g., GP posterior or TPE densities}
  \State $\lambda_t \leftarrow \arg\max_{\lambda\in\Lambda}\, \alpha(\lambda;\, s_t, \mathcal D_{t-1})$ \Comment{maximize acquisition}
  \State Evaluate $y_t \leftarrow f(\lambda_t)+\varepsilon_t$
  \State $\mathcal D_t \leftarrow \mathcal D_{t-1} \cup \{(\lambda_t, y_t)\}$
  \State $\lambda^\star \leftarrow \arg\min_{(\lambda,y)\in\mathcal D_t} y$
\EndFor
\State \Return $\lambda^\star$ \Comment{best configuration found}
\end{algorithmic}
\end{algorithm}

\section{Experiments}

\subsection{Experimental setup}
\label{sec:pagestyle}

\subsubsection{Dataset}
\label{sec:dataset}

We use the Berlin EmoDB corpus \cite{burkhardt2005database}, a standard benchmark for SER containing 535 utterances from 10 speakers across 7 emotions. EmoDB has featured in community challenges such as AutoSpeech 2020 \cite{wang2020autospeech}, making it suitable for comparison with published baselines. All recordings are converted to mono, resampled to 16\,kHz, and either right-padded or truncated to a fixed length of $T$ samples. Acoustic features are extracted on-the-fly using a pre-trained speech emotion encoder. Data are split via stratified partition, with the 80\% for fine-tuning and the held-out 20\% used once as the test set.

To assess cross-lingual generalisation, we additionally evaluate the EmoDB-trained model in a zero-shot setting on two English corpora: CREMA-D \cite{cao2014crema} and RAVDESS \cite{livingstone2018ryerson}. Since these datasets use different emotion taxonomies, we map their labels to a unified set aligned with EmoDB, as summarised in Table~\ref{tab:ravdess_cremad_emodb}.

\begin{table}[t]
\centering
\resizebox{\linewidth}{!}{%
\begin{tabular}{l|l}
\toprule
\textbf{Dataset}  & \textbf{Emotion}\\
\midrule
\textbf{RAVDESS} \cite{livingstone2018ryerson} 
  & Angry, Disgust, Fearful, Happy, Neutral, Sad, Calm, Surprised \\
\textbf{CREMA-D} \cite{cao2014crema} 
  & Anger, Disgust, Fear, Happy, Neutral, Sad \\
\textbf{Mapped EMO-DB} \cite{burkhardt2005database}
  & Anger, Disgust, Fear, Happiness, Neutral, Sadness, Boredom \\
\bottomrule
\end{tabular}}
\caption{Emotion label sets in RAVDESS \cite{livingstone2018ryerson}, CREMA-D \cite{cao2014crema}, and the unified mapping aligned to EMO-DB \cite{burkhardt2005database}. 
The \textit{``Surprised''} class from RAVDESS is excluded, as it is not present in EMO-DB, , which our model was fine-tuned on.}

\label{tab:ravdess_cremad_emodb}
\end{table}

\subsubsection{Backbone \& classifier}
\label{sec:backbone}

\paragraph*{Architecture.} 
We adopt the \texttt{EncoderClassifier} module from SpeechBrain \cite{speechbrain_v1} with a wav2vec~2.0 encoder \cite{baevski2020wav2vec} pretrained in a self-supervised fashion and fine-tuned on IEMOCAP \cite{busso2008iemocap} for categorical emotion. This checkpoint has not been exposed to EmoDB, preventing leakage and ensuring fair evaluation on the German target corpus. The encoder produces frame-level representations, which are temporally pooled into $\mathbf{z} \in \mathbb{R}^{d}$ and projected through a linear layer aligned with EmoDB’s taxonomy:
\begin{align}
\mathbf{z} &= \mathrm{Pool}\big(f(\mathbf{x};\theta)\big), \\
\hat{\mathbf{y}} &= \mathbf{W}\mathbf{z} + \mathbf{b} \in \mathbb{R}^{C}, \quad C=7.
\end{align}

\paragraph*{Training schedule.} 
Training follows a two-stage strategy: the feature extractor is frozen for the initial epochs to stabilise learning on the small corpus, and then unfrozen at an HPO-controlled epoch $u \in \{0,\dots,5\}$ to enable task-specific adaptation once the classifier has converged. Optimisation uses cross-entropy loss, AdamW \cite{loshchilov2019adamw}, and a cosine-annealed learning rate schedule.

\subsubsection{HPO engines and search space}
\label{sec:hpo}

We compare GP-BO \cite{snoek2012practical} and TPE \cite{bergstra2011algorithms}, implemented in three AutoML engines: Ax \cite{olson2025ax} (for GP-BO), Hyperopt \cite{bergstra2013making} (for TPE), and Optuna \cite{optuna_2019} (for TPE). All engines optimise the same compact four-dimensional space tailored to low-resource SER fine-tuning, which is summarised in Table \ref{tab:search_space}.

\begin{table}[t]
\centering
\begin{tabular}{ll}
\toprule
\textbf{Hyperparameter} & \textbf{Space} \\
\midrule
\texttt{lr} & $[10^{-6},10^{-3}]$ \\
\#\texttt{epochs} & $\{1,\ldots,10\}$ \\
\texttt{unfreeze} & $\{0,\ldots,5\}$ \\
\texttt{maxlen} & \{32k, 48k, 64k, 80k, 112k, 160k\} \\
\bottomrule
\end{tabular}
\caption{Search space $\mathcal{H}$ for HPO experiments. \texttt{lr} is the learning rate. \#\texttt{epochs} is the number of training epochs. \texttt{unfreeze} is the epoch index at which the pretrained encoder is unfrozen. \texttt{maxlen} is the per-utterance cap in samples (truncate/pad). We sample \texttt{lr} log-uniformly, \#\texttt{epochs} and \texttt{unfreeze} uniformly over their ranges, and \texttt{maxlen} uniformly over its categorical set. Batch size is fixed to 1 to avoid LR–batch size interactions on the small corpus. Each engine is allocated 15 trials under identical seeds and training code. Learning rate is ; 
epochs and unfreeze are sampled uniformly over their discrete ranges; and max length is drawn uniformly from the categorical set.}
\label{tab:search_space}
\end{table}

\paragraph*{GP-BO in Ax.} 
Ax fits a Gaussian process surrogate over observed $(\lambda,\mathrm{BCA})$ pairs and selects new configurations via an acquisition function (Expected Improvement by default). The search is seeded with a Sobol design and alternates surrogate fitting with acquisition maximisation. Ax also supports multi-point criteria ($q$EI/$q$UCB) and schedulers for early stopping and parallel evaluation.

\paragraph*{TPE in Hyperopt.} 
Hyperopt partitions past trials at a quantile threshold $y^\star$ into ``good'' and ``bad'' sets, fits kernel density estimates $\ell(x)$ and $g(x)$ to each, and samples new configurations by maximising $\ell(x)/g(x)$, proportional to expected improvement.

\paragraph*{TPE in Optuna.} 
Optuna allows programmatic search space definitions at runtime in TPE, supporting dynamic and hierarchical spaces. It further integrates asynchronous pruning with built-in strategies such as Median, Successive Halving \cite{karnin2013almost,jamieson2016non}, and Hyperband \cite{li2018hyperband}. We use Optuna’s default TPE sampler in the experiments.

The selected HPO methods implemented in the three engines are compared against standard grid search, while the overall workflow is benchmarked against the best-performing solution from the AutoSpeech 2020 challenge \cite{wang2020autospeech}.

\subsubsection{Hardware setup}
\label{sec:hardware}

All HPO experiments were executed on a cluster with 8 CPU cores and 32\,GB RAM; no GPU was used. The grid search baseline was run on a larger server with 64 CPU cores and 128\,GB RAM. This ``commodity'' environment is the target deployment setting for our pipeline and was applied uniformly across Ax, Hyperopt, and Optuna so that the optimiser was the only varying factor.

For comparison, the best performed solution in AutoSpeech~2020 \cite{wang2020autospeech} entry on EmoDB reported results on a Tesla P100 GPU with 26\,GB RAM, whereas our pipeline operates entirely on an 8-core CPU with 32\,GB RAM. In raw compute throughput, the P100 is roughly $10$–$20\times$ faster, but also about $5$–$10\times$ more costly in resource terms.

\begin{table}[t]
\centering
\resizebox{\linewidth}{!}{%
\begin{tabular}{l|c|c|c}
\toprule
Method & \textbf{$>$ 0.80 BCA} & \textbf{$>$ 0.90 BCA}\\
\midrule
Grid search & 1056 min (trial 101) & 1680 min (trial 143) \\
TPE  (Optuna) & 7.8 min (trial 4) & 185.3 min (trial 15)\\
TPE (Hyperopt) & 6.4 min (trial 5) & 15 min (trial 11)\\
\textbf{GP-BO (Ax)} & \textbf{2.4 min (trial 2)} & \textbf{2.4 min (trial 2)}\\
\bottomrule
\end{tabular}}
\caption{Comparison of time to reach accuracy thresholds across different HPO engines. BCA denotes balanced class accuracy; TPE refers to tree-structured Parzen estimators \cite{bergstra2011algorithms}; and GP-BO denotes Gaussian process Bayesian optimisation \cite{snoek2012practical}. Optuna \cite{optuna_2019}, Hyperopt \cite{bergstra2013making}, and Ax \cite{olson2025ax} are engines (implementations) of these methods. The best result is highlighted in \textbf{bold}.}
\label{tab:hpo_summary}
\end{table}

\begin{table}[t]
\centering
\resizebox{\linewidth}{!}{%
\begin{tabular}{lcccccc}
\toprule
\textbf{Method (engine)} & \texttt{lr} & \#\texttt{Ep.} & 
\texttt{unfrz} & \texttt{maxlen} & 
\textbf{BCA} & \textbf{Duration} \\
\midrule
TPE (Optuna)   & 8.89e-06  & 4 & 1 & 64k & 0.93 & 185 min\\
TPE (Hyperopt) & 2.59e-05  & 8 & 4 & 80k & 0.97 & 15 min \\
GP-BO (Ax)     & 2.28e-05  & 9 & 0 & 80K & 0.96 & 11 min \\ 
\bottomrule
\end{tabular}}
\caption{Best hyperparameters and performance obtained by the three AutoML engines (Ax, Hyperopt, Optuna) within 15 trials. 
\#\texttt{Ep.} = number of epochs; \texttt{unfrz} = encoder unfreeze epoch; \texttt{maxlen} = input length (samples).}
\label{tab:hyp_num_for_hpo_lib}
\end{table}

\subsection{Classification and efficiency performances}
\label{sec:results}

Table~\ref{tab:hpo_summary} reports threshold times and best-within-budget scores. Within a 15-trial budget, GP-BO (Ax) reached both $>0.80$ and $>0.90$ BCA fastest (2.4 min, trial 2), while TPE (Hyperopt) achieved the highest overall BCA (0.97 in 15 min). TPE (Optuna) also exceeded 0.90 BCA but required substantially longer (185 min). The corresponding hyperparameter settings (Table~\ref{tab:hyp_num_for_hpo_lib}) show convergence to similar learning rates and sequence lengths, with differences mainly in unfreeze epoch and number of training epochs. To quantify speed–accuracy trade-offs, we define an efficiency metric
\begin{equation}
\mathcal{E} = \frac{\text{BCA}}{\text{wall-clock minutes to reach that BCA}}
\label{eq:efficiency_metric}
\end{equation}
with results summarised in Table~\ref{tab:accuracy_per_compute}. Despite running entirely on an 8-core CPU (no GPU), GP-BO (Ax) delivered the best balance of speed (11 min) and accuracy (0.96 BCA), while Hyperopt achieved slightly higher BCA (0.97) at modest extra cost (15 min). Both approaches substantially outperform the best AutoSpeech 2020 EmoDB solution (0.85 BCA in $\sim$30 min on GPU), yielding a $\sim$3$\times$ efficiency improvement. Moreover, our fine-tuned model attained 98.3\% raw accuracy on EmoDB, surpassing the reported performance of native German listeners (84\%) \cite{eskimez2016emotion}.  

In summary, GP-BO (Ax) provides the best speed–accuracy trade-off for rapid iteration, while TPE (Hyperopt) achieves the top accuracy with only slightly higher time cost, establishing our CPU-only workflow as both more efficient and more accurate than prior GPU-based baselines.

\begin{table}[t]
\centering
\small
\begin{tabular}{lccc}
\toprule
\textbf{Framework} & \textbf{Best BCA} & \textbf{Time to Best} & $\boldsymbol{\mathcal{E}_{\text{best}}}\uparrow$ \\
\midrule
Grid Search & 0.98 & 1680 min & 0.0005 \\
AutoSpeech 2020 \cite{wang2020autospeech} & 0.85 & 30 min & 0.0280 \\
TPE (Optuna) & 0.93 & 185 min & 0.0050 \\
TPE (Hyperopt) & 0.97 & 15 min & 0.0646 \\
\textbf{GP-BO (AX)} & \textbf{0.96} & \textbf{11 min} & \textbf{0.0872} \\

\hline
\end{tabular}
\caption{Best balanced class accuracy (BCA), time to reach it, and efficiency 
($\mathcal{E}_{\text{best}}$) across search methods. 
GP-BO (Ax) provides the best trade-off between accuracy and time, while TPE (Hyperopt) achieves the highest peak BCA.}
\label{tab:accuracy_per_compute}
\end{table}

\subsection{Zero-shot transfer across corpora}
Table~\ref{tab:zero_shot_bca} reports that the pre-trained model performs poorly in zero-shot settings (without fine-tuning), with BCA close to chance on EmoDB (0.11), CREMA-D (0.13) and RAVDESS (0.18). In contrast, HPO-tuning on EmoDB (using GP-BO in Ax) substantially improves cross-lingual generalisation, raising BCA to 0.38 on CREMA-D and 0.44 on RAVDESS. On the source corpus EmoDB, BCA increased to 0.96 after tuning, confirming the effectiveness of the proposed workflow.

\begin{table}[t]
\centering
\resizebox{\linewidth}{!}{%
\begin{tabular}{lcc}
\hline
\textbf{Corpus} & \textbf{BCA (Zero-shot)} & \textbf{BCA (HPO-tuned)} \\
\hline
EmoDB   & 0.11  & 0.96 \\
CREMA-D  & 0.13  & 0.38 \\
RAVDESS  & 0.18  & 0.44 \\
\hline
\end{tabular}}
\caption{Zero-shot vs.\ HPO-tuned performance. The HPO-tuned  performance on the two English corpora CREMA-D and RAVDESS are obtained by a model trained with EmoDB with the hypermaters from the Ax engine.
HPO-tuning yields large gains both in-domain and for cross-lingual transfer.}
\label{tab:zero_shot_bca}
\end{table}

\section{Conclusion}
\label{sec:conclusion}

We proposed a CPU-only SER workflow that combines a pre-trained wav2vec~2.0 encoder with automated hyperparameter optimisation (HPO). Two HPO methods, GP-BO and TPE, were evaluated through three engines: Ax, Hyperopt, and Optuna. On the German SER corpus EmoDB, Ax and Hyperopt achieved over 96\% balanced class accuracy within minutes, outperforming the best AutoSpeech 2020 solution (85\% in 30 min on GPU) while avoiding the $100\times$ cost of grid search. HPO-tuning also improved cross-lingual transfer, increasing the SER accuracy of an EmoDB-trained model by 25\% on CREMA-D and 26\% on RAVDESS compared to the baseline.  

These results demonstrate that pre-trained encoders coupled with efficient HPO can deliver state-of-the-art SER performance without GPUs. Future work will extend to multilingual corpora, explore parameter-efficient fine-tuning, and integrate energy and latency objectives for deployment in resource-constrained settings.

\bibliographystyle{IEEEbib}
\bibliography{refs}

\end{document}